\documentclass[runningheads]{llncs}
\usepackage{graphicx}
\begin{document}
\title{The Artificial Scientist:\\ Logicist, Emergentist, and Universalist Approaches to Artificial General Intelligence\thanks{This work was supported by JST (JPMJMS2033; JPMJPR17G9).}}

\titlerunning{The Artificial Scientist}
%
\author{
Michael Timothy Bennett \and Yoshihiro Maruyama
}
\authorrunning{M. T. Bennett and Y. Maruyama}
%
\institute{School of Computing, Australian National University, Canberra, Australia\\
\email{michael.bennett@anu.edu.au} and
\email{yoshihiro.maruyama@anu.edu.au}}
\maketitle              
\begin{abstract}
We attempt to define what is necessary to construct an Artificial Scientist, explore and evaluate several approaches to artificial general intelligence (AGI) which may facilitate this, conclude that a unified or hybrid approach is necessary and explore two theories that satisfy this requirement to some degree. 

\end{abstract}
\section{Introduction}
Among the proposed means of verifying AGI, Goertzel's 2014 survey \cite{goertzel_2014} listed The Artificial Scientist Test \cite{adams_2012}, which stipulated that AGI will have been achieved when an artificial intelligence (AI) independently produces research sufficient to win a Nobel prize. While there is a wealth of research on AGI in general, and the automation of science has been explored to some extent \cite{king_2009}, what would be required to satisfy this test remains unclear. This paper attempts to clarify what exactly is necessary to create an Artificial Scientist, and how this fits within existing approaches to AGI.

A scientist may be many things, but for our purposes a simple and unambiguous definition is best. The Royal Society's motto, ``nullius in verba", serves nicely. Translated as ``take nobody's word for it", it emphasises that a scientist establishes truth through experiment, not testimony. For this, our agent must possess certain qualities.

\section{What is Required of an Artificial Scientist?}
This is not a list of every quality an Artificial Scientist ought to posses, but an attempt to identify what is necessary. 

\paragraph{\bf Representation of Hypotheses:} We'll define a hypothesis as a statement which has a truth value. A subset of such statements are readily testable, suitable subjects of scientific enquiry. We'll not concern ourselves with the specific language used to represent a hypothesis beyond stating that an Artificial Scientist must possess a means of representing any particular hypothesis.

\paragraph{\bf Inductive Inference:} The Royal Society's motto is an explicit rejection of testimony as the basis of any claim. An Artificial Scientist must not rely on testimony.  
Without testimony, what is true must be inferred through observation, and so the ability to perform inductive inference seems necessary. 

\paragraph{\bf Deductive and Abductive Reasoning:} Having inferred something of what \textit{is} an agent may transform this information, without speculation, through deductive reasoning. Then, from what is, our agent could abduct all that may be true, but uncertain. A testable hypothesis is one such thing, abducted from what is known. 
It seems necessary then for our agent to engage in deductive and abductive reasoning.

\paragraph{\bf Causal Reasoning and Explainability:} The purpose of an experiment is to test a hypothesis, identifying cause and effect \cite{pearl_2009}. Arguably this is desired in order that humans may develop the technology to reproduce or prevent that effect at will. A provisionally accepted hypothesis explains phenomena.
An explanation is only useful if it can be understood by its intended audience, and so a scientist must be able to communicate their hypotheses and the significance of their results in terms of what its audience values and understands. Mere interpretability is insufficient for more complex phenomena, as interpreting even simple symbolic models of well understood subject matter requires a great deal of technical expertise. Pushing forward the boundaries of scientific achievement would produce models of such complexity as to be beyond the capabilities of human interpretation. Fluency in natural language is desirable. 

\paragraph{\bf Evaluation of Hypotheses:} A hypothesis must at least be falsifiable, positing cause and effect. If we assume computational resources are finite, then there is a cost to consider in the search for hypotheses. The question then is which hypotheses \textit{ought} to be abducted. Hume's Guillotine tells us one cannot derive an ought from an is, and so we must give our agent an ought by which to judge hypotheses.
If one is to choose between several hypotheses, the truth of any one of which would serve to explain observed phenomena, then it seems reasonable to assert that one should start by testing the most plausible, the most likely to be true.
One must also consider what is gained by proving or disproving any hypothesis. Yes, one may choose to investigate with scientific rigour problems of no interest to anyone, but we would hesitate to claim this is accepted practice for contemporary scientists. 
Hence we assert that an Artificial Scientist must have a means of judging the plausibility of, and potential profit in any line of inquiry; a heuristic to inform its search of the space of possible hypotheses.

\paragraph{\bf Experimental Design, Evaluation and Planning:} To test a hypothesis one must design an experiment that isolates and tests the hypothesised cause of an effect, ideally controlling for all other variables. Each experiment costs resources, and the information gained should be evaluated in terms of expected benefit across hypotheses and future experiments. For example, a valuable experiment may not entirely confirm or disprove any one hypothesis, but may provide information allowing an agent to more efficiently select future experiments that will confirm or disprove many high priority hypotheses.
There are also risks to consider in an experiment. An experiment with a high expected utility may threaten the agent's continued existence, and so some form of risk aversion may be necessary (for example, when planning future experiments the geometric mean may be more appropriate than the arithmetic mean when computing utility, because the utility of future experiments depends upon the outcome of preceding experiments and their impact on available resources and capability).
An agent must identify what novel information would confirm or disprove those abducted hypotheses of the greatest expected utility.
It must design experiments that will convey said novel information and compare and plan experiments based on opportunity cost and risk.

\paragraph{\bf Enactivism:} To perform experiments, an agent must possess a means of interacting with the environment. The process of experimentation could be perceived as enactive cognition \cite{thompson_2007}, which posits cognition arises through the interaction of an organism with its environment. It assumes cognition is embodied, embedded to function within the confines of a specific environment, enacted through what an organism does and, finally, extending into that environment to store and retrieve information. All of this seems obviously necessary to conduct experiments in the environment. We are not offering an unqualified endorsement of embodied cognition; after all it is arguable that even a laptop has a body \cite{wang_2009}. However, experimentation has certain physical requirements, and if one is able to perform targeted experiments that obtain specific novel information and isolate causal relations, the process of learning may proceed much faster than if one is forced to wait until that same novel information is observed by chance.

\section{Three Relevant Approaches to AGI}
For the following we draw heavily upon Goertzel's 2014 survey of the field \cite{goertzel_2014}, deviating slightly to include recent developments and adjust the categories to suit our purposes. To standardise the terms with which we compare these approaches we employ a model of an arbitrary task, we treat the application of intelligence as prediction, and so define each of these approaches as trying to predict the appropriate response $r$ given a situation $s$. For the sake of brevity appropriateness, situation and response can be read using their common language definitions, but more concrete definitions are available if the reader is curious \cite{bennett_2021b,bennett_2021}.

\paragraph{\bf Logicist:}
We'll use the term logicist as a catch all for approaches that employ symbolic knowledge representation and inductive logic programming for learning. A finite set of symbols are used to describe discrete environment states, actions and so forth.
Symbols may be joined by logical connectives to specify statements that have a truth value within any given state. Degrees of belief in the truth of a statement, probabilities, may be assigned to statements where the environment is stochastic or partially observable.
Statements may also be employed as constraints to define what behaviour is permitted; the rules of a task. An agent may infer such constraints from examples through what is called inductive logic programming. 
As with Goertzel's symbolic category \cite{goertzel_2014} a logicist approach typically subscribes to The Physical Symbol System Hypothesis \cite{newell_1990,nilsson_2007}, meaning the abstract symbols it employed are assumed to be grounded in hardware, but how is a matter left for implementation. We go a step further and say that a logicist approach is perhaps better characterised as employing a constant and unchanging vocabulary of symbols which are in some manner specified by a human. Such an agent typically learns rules that determine whether a chosen response $r$ is correct in a situation $s$, from which a correct response $r$ may then be derived for a given a new situation $s$, allowing the agent to generalise. This is as opposed to modelling $r$ as a function of $s$ directly. As a result the choice of response is technically interpretable, but only in a limited sense, because the meaning of symbols is dyadic, exact, and parasitic on the meanings in the head of the human interpreter \cite{harnad_1990}. This is as opposed to emergent natural language, in which meaning is fluid and open to interpretation.

\paragraph{\bf Emergentist:}
Emergentist approaches take as their premise that complex behaviour and what we call abstract symbol systems may emerge through subsymbolic processes, such as the interaction of neurons. 
This process is called symbol emergence \cite{taniguchi_2019}, and typically uses approaches such as latent dirichlet allocation to cluster multi-modal sensorimotor stimuli into perceptual \cite{barsalou_1999} or sensorimotor \cite{milkowski_2018} symbols with fluid definitions akin to natural language. For example, researchers created an agent able to associate the sound of the word ``cup" with the image and other characteristics of a cup as experienced by that agent \cite{taniguchi_2016}, but success in constructing more complex symbols such as ``opera" or ``belief" remains elusive. 
Such an emergent symbol system could be used in conjunction with a logicist approach, which could then learn the rules of a given task in terms of these symbols. However, we extend this category to encompass approaches that bypass the construction of an abstract symbol system entirely in favor of directly modelling correct responses as a function of situations. For example, a neural network performing image classification. This is assumed to implicitly model the rules that determine the correctness of responses, but as a result is not as readily interpretable as logicist methods. 

\paragraph{\bf Universalist:}
In the context of an arbitrary task, an agent must map situations to responses. Such an agent could be conceived of as a program. A universalist approach assumes quite reasonably that for each environment (or task) there exists at least one program that always chooses the best possible responses (maximising reward), and so there must exist a program that maximises reward across all environments (or tasks).
To define this program Hutter \cite{hutter_2005} employed a formalisation of Ockham's Razor named Solomonoff's Universal Prior \cite{solomonoff_1964a,solomonoff_1964b} which assigns a weight to every program which reconstructs what the agent has experienced of the environment thus far. The same model could be expressed by programs of varying length, and so each program is evaluated by its Kolmogorov complexity \cite{kolmogorov_1963}; the smallest self extracting archive in a specific language. This theoretical agent, named AIXI, has been proven to perform such that there is no other agent which outperforms it in one environment that can also equal its performance in all others (in other words, AIXI may be outperformed in a specific environment by a more specialised agent employing inductive bias that prevents it performing as well across at least some other environments). While such an approach encompasses both logicist and emergentist approaches because it searches the space of all programs, it is distinct in that it begins with a guarantee of optimal performance in terms of an arbitrary reward function. The downside; Solomonoff Induction is incomputable. However, working approximations of AIXI have been constructed and this theoretical model provides useful insights about the significance of compression for intelligence.

\section{Are Any of These Approaches Alone Sufficient?}

In the following we discuss whether any of these is sufficient or not.

\subsection{Universalist}
\paragraph{\bf For:} By choosing the smallest self extracting archive as its model of the world, AIXI is in effect choosing from among possible hypotheses the most plausible according to what it has experienced so far. Certainly AIXI is capable of inference, and in extracting predictions from its highly compressed representation of the environment it seems reasonable to assume there must be something analogous to deduction or abduction taking place. Finally and most importantly, by finding the most compressed representation it must isolate causal relations \cite{budhathoki_2018}.

\paragraph{\bf Against:} Assuming AIXI could be approximated well enough, behaviours such as explainability must all somehow be specified by the reward function. Creating a function that guarantees such complex behaviour may not be any more achievable than AGI in general.

\subsection{Emergentist}
\paragraph{\bf For:} The utility, or at least popularity, of emergentist methods such as deep learning in narrow industrial applications seems almost indisputable. Models such as GPT-3 demonstrate that even complex writing tasks are not beyond reach with existing technology. While such models tend to be difficult to interpret, if they can be made to infer the ambiguous rules underlying natural language then perhaps they can eventually be made explainable to a layman. Certainly an emergentist method is easily implemented in a physical robot, because there is no abstraction required. If human language in all its inconsistency is to be acquired by an AI, then emergentist methods seem a promising approach. 

\paragraph{\bf Against:} An agent that mimics plausible explanations is of no more practical use as a scientist than an agent that gives no explanation at all. We can only trust explanations as far as we can interpret and verify them. Further, those explanations may be closer to mimicry than reasonable hypotheses. For example, the aforementioned GPT-3 seemed to acquire basic arithmetic, but as soon it was presented with less common sums it started giving responses that were wrong \cite{floridi_2020}. It appears to mimic arithmetic whilst having failed to grasp its rules. This may also be why many popular emergentist methods require so much data to learn in comparison to existing logistic or universalist methods; an agent that only mimics must learn all correct responses by rote, while an agent that understands what determines the correctness of all responses will be equipped to identify the correct one in any situation \cite{bennett_2021b}. All of these issues must be addressed before emergentist methods alone might result in an Artificial Scientist.

\subsection{Logicist}
\paragraph{\bf For:} A hypothesis represented as a statement within a predefined symbol system is interpretable. Though SAT is NP-Hard, a suite of existing solvers allows one to search the space of possible hypothesis fairly efficiently and with guaranteed optimality. Constraints are easily specified in comparison to other methods, allowing one to tailor agent behaviour to better suite specific tasks such as experimental design and planning. Finally, not only is the technology to deduct, abduct and infer with symbolic representations mature, but causal relations and their computation are typically defined in terms of symbols, and remain easily verifiable after the fact. 

\paragraph{\bf Against:} Abstract symbols must somehow connect to low level sensorimotor stimuli \cite{harnad_1990}. Even if this is solved, a fixed set of symbols chosen by a human may be far from suitable to express explanations for which we require an Artificial Scientist. Even if it were, such explanations are unlikely to be understood by even the most qualified of humans \cite{bennett_2021b,evans_2021b}. What is less obvious about cognition is why symbols are formed as they are. Which abstractions are best? This may be the most significant aspect of symbol emergence, that what emerges is part of the solution to a task, expressing specifically those things of relevance in solving it \cite{bennett_2021b}. What of extending cognition into parts of the environment never conceived of in the specification of the symbol system? While logicist methods may surpass emergentist in terms of interpretability, causal reasoning, data-efficiency and our ability to control, emergentist methods remain the state of the art by a large margin in terms of computer vision, natural language processing and so on. More, the simple act of representing a hypothesis symbolically does not mean it is the most accurate hypothesis explaining the data. Something akin to the formalisation of Ockham's Razor employed in universalist methods remains necessary.

\section{A Unifying Perspective}
None of the above appear sufficient in isolation, at least in the near term, for the purpose of constructing an Artificial Scientist. However, together they address all characteristics we deemed necessary. A universalist approach reveals what hypotheses are most plausible \cite{hutter_2005,legg_2008,chaitin_2006} and, by virtue of optimal lossless compression, isolate causal relations \cite{budhathoki_2018}. A logicist approach facilitates interpretable representation of hypotheses, planning, causal reasoning \cite{pearl_2018,pearl_2009,evans_2021a,evans_2020,evans_2021b,bennett_2021,bennett_2021b} and the ability to tailor behaviour to our needs with ease. An emergentist approach facilitates enactivism and the possibility of an emergent symbol system which is efficient \cite{bennett_2021b,bennett_2021}, fluid and comparable to natural language \cite{taniguchi_2016,taniguchi_2019,milkowski_2018}.
Two complimentary bodies of research may provide a foundation for future work along these lines; the formalisation of an arbitrary task and its solutions (named ``The Solutions to Any Task") \cite{bennett_2021b}, and a formalisation inspired by the work of Kant \cite{evans_2020} (named ``Kant's Cognitive Architecture"). These approaches are similar but based on different premises, providing different insights. We will now briefly summarise and compare them as they pertain to developing an Artificial Scientist.
Both attempt to infer a hypothesis which explains observed data, from which correct responses to every situation may be abducted. Combined with a SAT solver to decode responses, such an hypothesis qualifies as lossless compression. Finally, neither approach relies upon abstract symbols, learning from sequences of sensory data in the case of Kant's Cognitive Architecture, and a set of sensorimotor sates in the case of The Solutions to Any Task. 

\paragraph{\bf Kant's Cognitive Architecture:}
Taking Kant's Critique of Pure Reason as its inspiration, this approach asserts that there is no such thing as a specific judgement. Subsequently every rule is universally quantified, ``doomed to generalise" as the authors put it \cite[p. 31]{evans_2021b}. Evans introduced notions of unity, a form of inductive bias that specifies which constraints are acceptable in terms of spacial, temporal and causal relations, along with object permanence (static unity). The solution is then made more general by choosing weak constraints. Such notions are well suited to explain all sensory data in general terms, and their implementation in the form of The Apperception Engine performs as one would expect (extremely well). The resulting hypotheses are general, perhaps not the most general, but enough that we can say that to some extent it accounts for the universalist's notion of plausibility (similar to Kolmogorov Complexity). It is arguably embodied to some extent, being concerned with sensory data, but does not account for the motor part of the sensorimotor system as is reflected by its specific form of inductive bias.

\paragraph{\bf The Solutions to Any Task:}
The notion of an arbitrary task attempts to formalise anything we might call a task in terms of its solutions (the assumption being that it must be possible to succeed or fail at a task to some degree, however the task need not be computable). In the domain of possible solutions to a given task there exist two extremes; an Intensional Solution (which may not be unique) and an Extensional Solution (which is unique). Because the task is defined in terms of a set of sensorimotor states, situation and response pairs rather than sequences of sensory data, the solution is more general, pertaining to interactive sensorimotor control rather than just the prediction of sensory input, with a much simpler inductive bias. The Intensional Solution is formed of the weakest, least specific rules necessary and sufficient to reconstruct the aforementioned set of sensorimotor states given the complete set of situations, a formalisation of Ockham's Razor which maximises the ability to generalise and identifies causal relations. The Extensional Solution is formed of the strongest, representing perfect mimicry with no generalisation. The Intensional Solution or a solution close to it represents intent, and is used to explain both symbol emergence \cite{bennett_2021b} and the modelling of intent in other agents \cite{bennett_2021}. The Intensional Solution is incomputable in general, but computable if restricted to a specific hardware language (a further inductive bias).

\paragraph{\bf Comparison with Respect to Hypotheses:}
These two approaches are not mutually exclusive, but complimentary. Given a subset of possible tasks pertaining to specific types of sensory input, the inductive bias implemented in the Apperception Engine will result in an Intensional Solution. For some other tasks, it may result in something more intensional than extensional, but for the remainder of tasks it may produce nothing useful (because it assumes ``there is no such thing as a specific judgement" \cite[p. 31]{evans_2021b}).
Consider a task to reproduce a set of random binary sequences, drawn from a uniform distribution, given only part of each sequence. The Apperception Engine would attempt to find what all sequences in the set share in common, and fail. There are no universal rules by which the sequences may be reproduced. In contrast, the Intensional Solution to the task would specify each sequence in detail, effectively rote-learning the set (the Intensional and Extensional Solutions would be one and the same). 
Ultimately, our Artificial Scientist should prefer hypotheses which are the most plausible regardless of task, meaning the Intensional Solution based on Ockham's Razor rather than the more restrictive inductive bias towards the aforementioned class of sensory sequences. However, a working implementation of the Apperception Engine is publicly available. It is not an abstract promise of future capability.

\paragraph{\bf Comparison with Respect to Experimentation:} The specifics of experimental design are not addressed in either case.
To illustrate why observational data alone may be insufficient, consider a task to either multiply two binary numbers, or add them. There are now two correct responses when presented with any two binary numbers. A set of situation-response pairs is observed, stipulating one and only one correct response for each situation. A response here is the result of either addition, or multiplication, the choice of which being made by a fair coin flip (uniformly distributed). This last part is important, as knowing a situation is observed with addition, or with multiplication, would not convey any useful information which would allow one to form two universally applicable mutually exclusive rules based on aspects of the situation. The Apperception Engine would fail because there is no single universal rule which is mutually exclusive with all others (there are in fact two concurrent rules). In contrast, the Intensional Solution would specify both correct rules, albeit restricted to specific rote memorised situations (it would be more specific than necessary because it would state that addition is necessary in some specific situations, and multiplication in all others). To find an Intensional Solution that properly describes the task (as two separate concurrent and universally applicable rules) would require the agent experiment, to test whether each rule holds in specific situations, or if both apply in all situations. An Intensional Solution alone is insufficient, because the information necessary to confirm that two mutually exclusive responses are valid in any given situation is not present in the existing set of observations. A hypothesis must be formed, abducted from the data, and tested to confirm whether the rules apply concurrently or in an alternating pattern.
As illustrated by the above task with two solutions, regardless of inductive bias experiment remains necessary to guarantee the most data-efficient mode of learning (to find the \textit{most} correct hypothesis).
\paragraph{\bf Comparison with Respect to Symbol Emergence and Explainability:} While Kant's Cognitive Architecture does not attempt to address symbol emergence, Evans has proposed the integration with subsymbolic methods such as neural networks to ground abstract symbols \cite{evans_2021b}, similar to the aforementioned work on symbol emergence in robotics \cite{taniguchi_2016}. These could then be composed into concepts by the Apperception Engine, albeit limited by the inductive bias towards sensory input. This would also seem to imply a constant and unchanging set of abstract symbols specified by a human (where the exact meaning of those symbols is determined by learning algorithm), but it is a step in the direction of an emergent symbol system.
However, The Solutions to Any Task posits that not all symbol systems are equal, that some are better suited to describe what is relevant to a task than others. As such a symbol system is implicit in the solution to a task, clustering sensorimotor stimuli in terms of what is relevant to success in that task \cite{bennett_2021b}. The distinction between any two symbols is then fluid, dependent upon context. This theory attempts to address not only the emergence of a symbol system, but the modelling of intent in other agents, empathy \cite{bennett_2021}, in aid of constructing explanations in natural language tailored to what the audience understands and considers important. To learn such a symbol system requires that symbols are not learned separately from one's model of the world, but as part of it so that they have meaning (``significance in terms of a goal") beyond their relation to other symbols. This is a fundamentally different approach to symbol emergence to the one proposed for the Apperception Engine, sharing in more in common with the notion of concepts discussed in Kant's Cognitive Architecture, but perhaps requiring a fundamentally different inductive bias to that of the Apperception Engine.


%
%

%
%
%

\begin{thebibliography}{8}

\bibitem{goertzel_2014}
Goertzel, B.: Artificial General Intelligence: Concept, State of the Art, and Future Prospects. In: Journal of Artificial General Intelligence 5(1), pp. 1--48 (2014)

\bibitem{adams_2012}
Adams, S. et al. Mapping the landscape of human-level artificial general intelligence. In: AI Magazine 33(1), pp. 25--42 (2012).

\bibitem{evans_2021a}
Evans, R., Hernández-Orallo, J., Welbl, J.,  Kohli, P., Sergot, M.:
Making sense of sensory input. In: Artificial Intelligence 293, (2021)

\bibitem{evans_2021b} Evans, R., Bošnjak, M., Buesing, L., Ellis, K., Pfau, D., Kohli, P., Sergot, M.: Making Sense of Raw Input. In: Artificial Intelligence 299, (2021)

\bibitem{evans_2020} Evans, R.: Kant's Cognitive Architecture. PhD Thesis, Imperial College London. (2020)

\bibitem{hutter_2005}
Hutter, M.: Universal Artificial Intelligence: Sequential Decisions based on Algorithmic Probability. Springer. (2005)

\bibitem{budhathoki_2018} Budhathoki, K., Vreeken, J.: Origo: Causal Inference by Compression. In: Knowledge and Information Systems 56(2), pp. 28--307 (2018)

\bibitem{chaitin_2006} Chaitin, G.: The Limits of Reason. In: Scientific American 294(3), pp. 74--81 (2006)

\bibitem{legg_2008} Legg, S.: Machine Super Intelligence. (2008)

\bibitem{taniguchi_2019} Taniguchi, T. et al.: Symbol Emergence in Cognitive Developmental Systems: A Survey. In: IEEE Transactions on Cognitive and Developmental Systems 11(4), pp. 494--516 (2019)

\bibitem{taniguchi_2016} Taniguchi, T. et al.: Symbol Emergence in Robotics: A Survey. In: Advanced Robotics 30(11-12), pp. 706--728 (2016)

\bibitem{chollet_2019} Chollet, F.: On the Measure of Intelligence. arXiv: 1911.01547[cs.AI], (2019)

\bibitem{kolmogorov_1963} Kolmogorov, A. N.: On tables of random numbers. In: Sankhya: The Indian Journal of Statistics A, pp. 369--376 (1963)

\bibitem{solomonoff_1964a} Solomonoff, R. J.: A formal theory of inductive inference. Part I. In: Information and Control 7(1), pp. 1--22 (1964)

\bibitem{solomonoff_1964b} Solomonoff, R. J.: A formal theory of inductive inference. Part II. In: Information and Control 7(2), pp. 224--254 (1964)

\bibitem{wang_2009} Wang, P. 2009. Embodiment: Does a Laptop Have a Body? In: Proceedings of AGI-09, pp. 74--179 (2009) 

\bibitem{bennett_2021} Bennett, M. T., Maruyama, Y.: Philosophical Specification of Empathetic Ethical Artificial Intelligence. To appear in: IEEE Transactions on Cognitive and Developmental Systems (2021)

\bibitem{bennett_2021b} Bennett, M. T.: Symbol Emergence and The Solutions to Any Task. 14th Conference on Artificial General Intelligence. (2021)

\bibitem{nilsson_2007} Nilsson, N. J.: The Physical Symbol System Hypothesis: Status and Prospects. In 50 Years of Artificial Intelligence. Springer. pp. 9--17 (2007)

\bibitem{pearl_2009} Pearl, J.: Causality: Models, Reasoning and Inference. 2nd. USA: Cambridge University Press. (2009)

\bibitem{pearl_2018} Pearl, J., Mackenzie, D.: The Book of Why: The New Science of Cause and Effect. 1st. USA: BasicBooks, Inc. (2018)

\bibitem{floridi_2020} Floridi, L., Chiriatti, M.: GPT-3: Its Nature, Scope, Limits, and Consequences. In: Minds and Machines pp. 1--14 (2020)

\bibitem{newell_1990} Newell, A.: Physical Symbol Systems. In: Cog. Sci., pp. 135--183 (1980)

\bibitem{harnad_1990} Harnad, S.: The Symbol Grounding Problem. (1990)

\bibitem{thompson_2007} Thompson, E.: Mind in Life. In: Biology, Phenomenology and the Sciences of Mind 18 (2007)

\bibitem{barsalou_1999} Barsalou, L. W.: Perceptual Symbol Systems. In: Behavioral and Brain Sciences 22(4), pp. 577--660 (1999)

\bibitem{milkowski_2018} Miłkowski, M.: Embodied Cognition. (2018) 

\bibitem{king_2009} King R. D. et al.: The Automation of Science. In: Science 324, pp. 85--89 (2009)








\end{thebibliography}
%

\end{document}